\documentclass[letterpaper]{article} 
\usepackage[preprint]{aaai2027}  
\usepackage[hyphens]{url}  
\usepackage{graphicx} 
\usepackage{natbib}  
\usepackage{caption} 
\usepackage{algorithm}
\usepackage{algorithmic}

\usepackage{newfloat}
\usepackage{listings}
\DeclareCaptionStyle{ruled}{labelfont=normalfont,labelsep=colon,strut=off} 
\floatstyle{ruled}
\newfloat{listing}{tb}{lst}{}
\floatname{listing}{Listing}

\usepackage{booktabs}

\usepackage{graphicx,verbatim}
\usepackage{multirow} 
\usepackage[dvipsnames]{xcolor}
\usepackage{amsfonts}
\usepackage{amsmath}
\usepackage{amssymb}
\usepackage{dsfont}

\usepackage{booktabs}

\usepackage{pifont}
\usepackage{kotex}

\newcommand{\eg}{{\textit{e.g.}}}      

\usepackage{amssymb}
\usepackage{amsmath}
\usepackage{booktabs}
\usepackage{multirow}
\usepackage{caption}
\usepackage{listings}

\title{STRIVE: Multi-Agent Structured Temporal Reasoning with Integrated Verification for Longitudinal Radiology Report Generation}

\author{
    Junyeong Maeng\textsuperscript{\rm 1},
    Eunsong Kang\textsuperscript{\rm 2}\corresponding,
    Heung-Il Suk\textsuperscript{\rm 1}\corresponding
}

\affiliations{
    \textsuperscript{\rm 1}Department of Artificial Intelligence, Korea University, Seoul, Republic of Korea\\
    \textsuperscript{\rm 2}Graduate School of Data Science, Kangwon National University, Chuncheon, Republic of Korea\\
    mjy8086@korea.ac.kr, eskang@kangwon.ac.kr, hisuk@korea.ac.kr
}

\begin{document}

\maketitle

\begin{abstract}

Longitudinal radiology report generation (LRRG) requires identifying both current findings and their changes relative to a prior study. Existing methods jointly model diagnosis, attribute estimation, temporal comparison, and language generation within implicit representations, which can cause task interference, obscure the evidence underlying each decision, and limit error traceability. They also model progression states as independent labels, ignoring their ordered structure and thus treating missed changes and direction reversals equally. We present STRIVE, Multi-Agent Structured Temporal Reasoning with Integrated Verification for LRRG, which decomposes clinical reasoning into specialized Diagnosis, Attribute, and Temporal Change Agents that produce explicit intermediate evidence. In particular, the Temporal Change Agent is further post-trained using Progression-Aware GRPO, a verifiable, shaped reward that assigns partial credit to direction-preserving errors while scoring direction reversals lowest. STRIVE performs verification at two stages: a deterministic Consistency Gate reconciles the agent outputs before report generation, and a Validation Agent checks whether the generated report is supported by the aggregated clinical evidence. On Longitudinal-MIMIC, STRIVE attains the best clinical efficacy among recent methods and more than doubles Longitudinal Change Concordance (LCC), a measure of temporal agreement with the reference report, over the strongest baseline.

\end{abstract}


\section{Introduction}

\begin{figure}[t]
  \centering
  \includegraphics[width=1\linewidth]{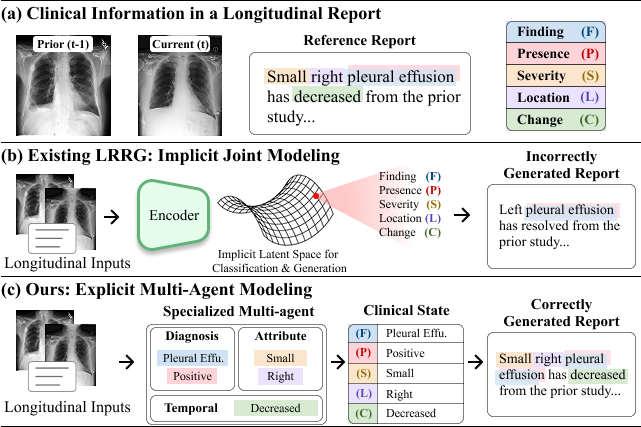}
  \caption{(a) A longitudinal report includes finding, presence, severity, location, and change. (b) Existing LRRG entangles them in an implicit latent space, generating clinically inconsistent reports. (c) STRIVE decomposes them across specialized agents, forms an explicit clinical state, and verifies the final report.}
  \label{fig1:compare}
\end{figure}

Radiology report generation (RRG) aims to automatically generate clinically meaningful reports from medical images. It is a challenging task because models must accurately identify fine-grained clinical information, including disease presence, severity, and location, and faithfully express it in natural language~\cite{brady2018radiology}. Recent advances in vision-language models and large language models have improved the linguistic fluency and clinical accuracy of generated reports, driving active research in RRG~\cite{wang2023r2gengpt,wang2024r2gencsr,liu2024medlmm}. More recently, this task has been extended to longitudinal radiology report generation (LRRG), which compares studies across multiple time points to capture disease progression beyond findings observed at a single time point~\cite{wang2024hergen,nicolson2024longitudinal,liu2026priorrg} as illustrated in Figure~\ref{fig1:compare}(a).

Recent LRRG methods incorporate longitudinal information in different ways. Alignment-based approaches establish temporal correspondences between prior and current studies by aligning visual or vision-language features~\cite{liu2026biotprompt,gao2026mare}, whereas fusion-based approaches integrate features across studies into a unified longitudinal representation for report generation~\cite{wang2024hergen,dong2026tim}, as illustrated in Figure~\ref{fig1:compare}(b). Despite effectively incorporating comparative information, these approaches face two key challenges: the implicit coupling of clinical reasoning and report generation, and inadequate modeling of directional relationships among progression states.

The first limitation stems from implicit coupling of diagnosis, clinical attribute estimation, temporal reasoning, and report generation within a shared representation. These tasks impose different representational demands. Report generation favors a smooth semantic space, whereas clinical objectives require a discrete and constrained decision space. Their joint optimization can cause task interference, obscure fine-grained evidence, and distort inferred information during generation~\cite{zhangdouble}. Moreover, when these reasoning processes are exposed only through the final report, it becomes difficult to determine whether an error originates from clinical reasoning or language generation. Explicit intermediate evidence is therefore essential for error localization, analysis, and clinical verification~\cite{tanida2023rgrg}.

The second limitation lies in inadequate modeling of relationships among progression states. Although recent LRRG methods use progression labels or change descriptions as supervision~\cite{yun2025diff,dong2026tim}, they typically model these states as independent targets. This overlooks their directional structure: \texttt{new} and \texttt{resolved} represent opposite transitions, while \texttt{increased}, \texttt{stable}, and \texttt{decreased} reflect ordered changes in persistent findings. Consequently, conventional objectives do not adequately distinguish a missed progression, such as predicting \texttt{stable} instead of \texttt{increased}, from a direction reversal, such as predicting \texttt{decreased} instead of \texttt{increased}, despite their different clinical and directional severity.

Together, these limitations motivate an explicit multi-agent framework~\cite{khan2026cograd,zhang2026radagents} that separates diagnosis, attribute estimation, and temporal reasoning across specialized agents, enabling task-specific optimization while making intermediate clinical evidence explicit. However, agent specialization may introduce cross-agent inconsistency, where agents produce conflicting findings, and evidence-to-report unfaithfulness, where valid intermediate evidence is omitted or distorted during report generation. Reliable multi-agent LRRG therefore requires integrated verification to keep the final report grounded in aggregated clinical evidence.

Alongside these modeling considerations, evaluation should also reflect the temporal correctness of generated reports. Existing diagnosis-based~\cite{smit2020chexbert}, lexical~\cite{papineni2002bleu,lin2004rouge,banerjee2005meteor}, model-based~\cite{zhangbertscore,ostmeier2024green}, and structure-based~\cite{jain1radgraph,yu2023evaluating} metrics capture complementary aspects of report quality, including disease accuracy, linguistic similarity, semantic agreement, and structural consistency. However, progression-specific properties are assessed only indirectly. A more direct evaluation of disease-specific change coverage and directional agreement is therefore needed for LRRG.

To address these limitations, we propose STRIVE: Multi-Agent Structured Temporal Reasoning with Integrated Verification for Longitudinal Radiology Report Generation. STRIVE decomposes longitudinal interpretation into specialized clinical reasoning agents, explicitly models progression-state relationships, and integrates evidence reconciliation with report-level validation as illustrated in Figure~\ref{fig1:compare}(c). Our contributions are fourfold:

\begin{itemize}
    \item We propose STRIVE, a multi-agent LRRG framework that decomposes diagnosis, clinical attribute estimation, and temporal progression modeling into specialized agents with explicit intermediate evidence.
    \item We introduce Progression-Aware GRPO, a reinforcement learning with verifiable rewards (RLVR) objective whose shaped, multi-level reward captures the directional and graded relationships among longitudinal progression states (\eg, increased, decreased, stable).
   \item We develop a Consistency Gate to resolve logical conflicts among agent outputs and a Validation Agent to verify that the final report is supported by the committed clinical state.
    \item We evaluate STRIVE on Longitudinal-MIMIC and show that it outperforms existing methods in linguistic fluency, diagnostic performance, and progression-state agreement measured by Longitudinal Change Concordance (LCC).

\end{itemize}

\begin{figure*}[t]
 \centering
 \includegraphics[width=\textwidth]{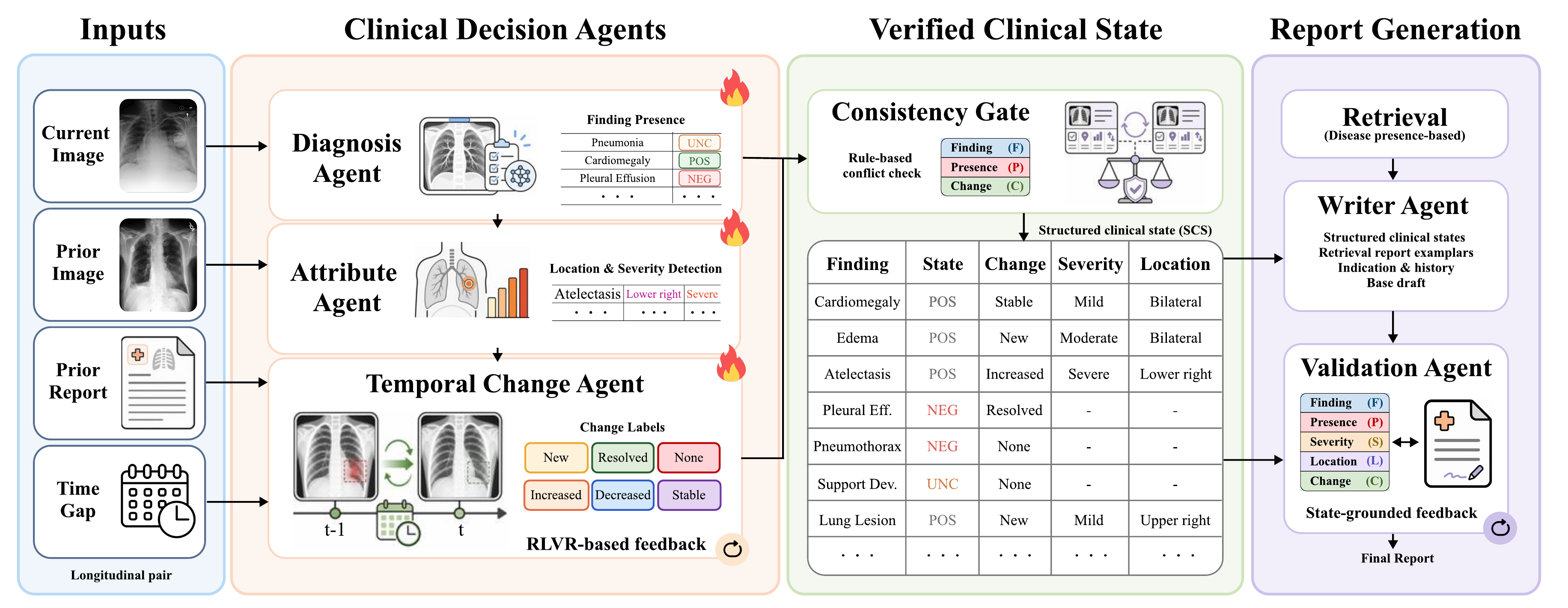}
 \caption{Overview of STRIVE. Three Clinical Decision Agents infer per-finding presence (Diagnosis), severity and location (Attribute), and change labels (Temporal Change) post-trained with GRPO. A rule-based Consistency Gate resolves diagnosis–change conflicts and aggregates the outputs into a structured clinical state. A Writer Agent then generates the report and a Validation Agent applies state-grounded edits to produce the final report.}
 \label{fig:framework}
\end{figure*}

\section{Related Work}
\label{sec:related_work}

\subsection{Longitudinal Radiology Report Generation} 
Conventional RRG primarily focuses on recognizing clinical findings, including disease presence, severity, and location, from a single study~\cite{chen2020r2gen,chen2021r2gencmn,wang2023r2gengpt,jin2024promptmrg}. However, methods based on a single study do not directly model cross-time disease changes, which require comparisons between prior and current studies. LRRG extends RRG by incorporating historical studies and modeling temporal evidence. 

Existing LRRG methods differ mainly in how they relate the prior study to the current one.
Fusion-based methods, such as PriorRG~\cite{liu2026priorrg}, integrate prior and current visual features into a unified spatiotemporal representation. Alignment-based methods explicitly establish cross-time relationships between local visual regions. BiOTPrompt~\cite{liu2026biotprompt} employs bidirectional optimal transport to identify asymmetric patch-level changes, whereas MARE~\cite{gao2026mare} dynamically aligns lesion regions and performs analogical reasoning over visual and textual evolution relations. TIM~\cite{dong2026tim} instead separates the spatial representation of each finding from the modeling of how it progresses between the two studies.

Despite their architectural diversity, existing LRRG methods encode disease recognition, clinical attribute estimation, and temporal comparison primarily within shared implicit representations. Such joint modeling causes interference among heterogeneous reasoning tasks, making errors difficult to trace and verify.

\subsection{Multi-Agent Clinical Reasoning and Verification}

Recent RRG studies have begun to adopt multi-agent formulations that structure image interpretation and report generation into specialized stages based on clinical workflows. For example, CogRad~\cite{khan2026cograd} assigns global triage, regional investigation, report writing, and output verification to Scout, Investigator, Writer, and Verifier agents, respectively. RadAgents~\cite{zhang2026radagents} divides chest X-ray interpretation by anatomical region and integrates the per-region analyses with a Synthesizer agent. Both also verify their outputs, CogRad by re-examining sentences that lack sufficient visual support and RadAgents by resolving inconsistencies across agent and tool outputs. These formulations, however, decompose the interpretation of a single study, so no agent is responsible for deciding how a finding has changed relative to a prior one. STRIVE makes that decision an explicit agent output and verifies it against the diagnosed presence before the report is written.

\section{Method}
\label{sec:method}

\subsection{Framework Overview}
\label{sec:overview}

Let $X_j=(I_j,v_j)$ denote the chest X-ray image $I_j$ and its acquisition view $v_j$ at time $j\in\{t-1,t\}$. The most recent prior study also provides its report $r_{t-1}$ and the inter-study interval
$\delta_t$. We use $c_t$ for the available clinical context, restricted to
the study history and indication. Our framework, STRIVE, generates the current
report $\hat r_t$ as
\begin{equation}
    \hat r_t=
    \mathcal G(X_t,X_{t-1},r_{t-1},\delta_t,c_t).
    \label{eq:overall}
\end{equation}
We apply task decomposition to longitudinal report generation, factorizing it into three clinical decisions: presence in the current study, attributes of a present finding, and change relative to the prior study. Each decision is handled by a role-specialized agent. The \textit{Diagnosis Agent} determines whether each finding is present, the \textit{Attribute Agent} characterizes present findings in terms of severity and location, and the \textit{Temporal Change Agent} identifies how each finding has changed since the prior study. We further perform verification at two stages. Before report generation, the \textit{Consistency Gate} corrects change states that are incompatible with the diagnosis state. After report generation, the \textit{Validation Agent} checks whether the generated report is supported by the structured clinical state. Figure~\ref{fig:framework} illustrates the overall pipeline.

\subsection{Clinical Decision Agents}
\subsubsection{Diagnosis Agent}
\label{sec:diagnosis_agent}
Let $\mathcal D$ denote the set of 14 CheXpert findings, consisting of 12 disease-related findings and two non-disease labels, $\{\textit{Support Devices}, \textit{No Finding}\}$.
The Diagnosis Agent independently estimates the state of every finding $d\in\mathcal D$ from the corresponding study. It receives complementary evidence from seven frozen pretrained chest X-ray experts, comprising three classification experts and four generative experts. The classification experts provide continuous finding scores, whereas the generative experts produce report-like textual evidence that is converted into categorical finding states by a CheXbert labeler~\cite{smit2020chexbert}.

Let $\mathbf e_j$ denote the resulting expert evidence at time $j$.
The expert evidence $\mathbf e_j$ and acquisition view $v_j$ are serialized in a fixed finding order and integrated by an instruction-tuned LLM $\mathcal A_D$:
\begin{equation}
\hat{\mathbf{s}}_j
=
\mathcal{A}_D(\mathbf{e}_j,v_j)
\in
\mathcal{Y}_D^{|\mathcal D|},
\end{equation}
where $\mathcal Y_D=\{\mathtt{POS},\mathtt{UNC},\mathtt{NEG}\}$ and
$\hat{s}_{j,d}$ denotes the predicted state of finding $d$ at time $j$.

\subsubsection{Attribute Agent}
\label{sec:attribute_agtent}

The Attribute Agent characterizes findings predicted as positive in the current study. 
For each such finding, a medical vision-language model $\mathcal A_A$ is prompted with the current chest X-ray $I_t$ and returns its severity and location, where location may include laterality and anatomical region. We apply the Attribute Agent to the 12 disease-related findings, denoted by $\mathcal D_A=\mathcal D\setminus\{\textit{Support Devices},\textit{No Finding}\}$.

For each $d\in\mathcal D_A$, the attribute output is defined as
\begin{equation}
\hat{\mathbf a}_{t,d}
=
\begin{cases}
\mathcal A_A(I_t,d),
& \hat s_{t,d}=\mathtt{POS},\\
\bot,
& \text{otherwise},
\end{cases}
\end{equation}
where $\mathcal A_A(I_t,d)=(\hat a^{\mathrm{sev}}_{t,d},\hat a^{\mathrm{loc}}_{t,d})$ gives the predicted severity and location, and $\bot$ indicates that no attribute prediction is available because the finding is not diagnosed as positive.

\subsubsection{Temporal Change Agent}
\label{sec:temporal_change}
The Temporal Change Agent $\mathcal A_T$ determines how each finding has changed from the prior to the current study. It operates on the same 12 disease-related findings as the Attribute Agent, and we therefore define $\mathcal D_T=\mathcal D_A$. The agent produces a change-state
\begin{equation}
\tilde{\mathbf m}_t
=
\mathcal A_T
\left(
r_{t-1},
\delta_t,
\hat{\mathbf s}_{t-1},
\hat{\mathbf s}_t,
\mathbf p_{t-1},
\mathbf p_t
\right),
\end{equation}
where $r_{t-1}$ is the prior report, $\delta_t$ is the inter-study interval, and $\hat{\mathbf s}_{t-1}$ and $\hat{\mathbf s}_t$ are the diagnosis states of the prior and current studies, respectively. The $\mathbf p_{t-1}$ and $\mathbf p_t$, produced by the three classification experts used in the Diagnosis Agent, contain the predicted presence probabilities for the 12 disease-related findings in the prior and current studies, respectively. Each element $\tilde m_{t,d}$ represents the change state of finding $d$ and takes one of six labels $\mathcal{Y}_T= \{ \mathtt{new}, \mathtt{increased}, \mathtt{stable}, \mathtt{decreased}, \mathtt{resolved}, \mathtt{none} \}$. The label $\mathtt{none}$ indicates that no longitudinal change state is assigned to the finding, so that a change state is defined for every finding in $\mathcal D_T$. Rather than directly comparing the raw image pair, the agent uses disease-specific presence probabilities and the prior report, providing an explicit and interpretable basis for finding-wise temporal reasoning.

\subsection{Training Strategies for Clinical Decision Agents}
\subsubsection{Supervised Fine-Tuning}

We first optimize each of the three agents using supervised fine-tuning (SFT). 
The Diagnosis Agent $\mathcal A_D$ is fine-tuned to predict the finding states in $\mathcal Y_D$, while the Attribute Agent $\mathcal A_A$ is trained to estimate the severity and location of positive findings. The attribute targets are extracted from the training reports using an LLM-based extractor. The Temporal Change Agent $\mathcal A_T$ is fine-tuned to predict disease-wise change states defined in $\mathcal Y_T$. To promote consistency across both temporal directions, we augment each valid training pair with a time-reversed counterpart. The prior and current evidence are swapped, along with the corresponding change labels: \texttt{new} and \texttt{resolved} are interchanged, as are \texttt{increased} and \texttt{decreased}, while \texttt{stable} and \texttt{none} remain unchanged. This augmentation encourages directionally consistent predictions when the study order is reversed.

\subsubsection{Progression-Aware GRPO for Temporal Change Agent}
We post-train the Temporal Change Agent $\mathcal A_T$ with a two-stage pipeline: SFT warmup followed by a reinforcement-learning (RL) stage~\cite{guo2025deepseek}. In the RL stage we optimize $\mathcal A_T$ with Group Relative Policy Optimization (GRPO)~\cite{shao2024deepseekmath}. Since the reward is computed programmatically from the structured change labels rather than by a learned reward model, our objective is an instance of RLVR. For each input, the policy samples a group of $G$ candidate responses $\{y_i\}_{i=1}^{G}$. Each response is assigned a reward $R_i=R(y_i,y^*)$ with respect to the report-derived target $y^*$. The policy is then updated using the standard clipped GRPO objective, in which the advantage of $y_i$ is its reward standardized within the group.

The reward is a shaped reward designed to reflect the structured relationships among longitudinal change states. A naive outcome (exact-match) reward treats all incorrect predictions equally, whereas our multi-level formulation assigns partial credit to direction-preserving errors. For example, predicting \texttt{increased} when the target is \texttt{new} preserves the worsening direction, whereas predicting \texttt{decreased} indicates the opposite direction. We therefore evaluate each response at three levels: detection, coarse-grained direction, and fine-grained change state.
\begin{equation}
\label{eq:temporal_reward}
R(y,y^*)
=
\frac{1}{3}
\left(
\mathrm{F1}_{\mathrm{detect}}
+
\mathrm{F1}_{\mathrm{coarse}}
+
\mathrm{F1}_{\mathrm{fine}}
\right).
\end{equation}
At the detection level, the five explicit change states are grouped together, while \texttt{none} indicates that no change state is assigned. At the coarse level, \texttt{new} and \texttt{increased} are grouped as worsening, \texttt{decreased} and \texttt{resolved} as improving, while \texttt{stable} and \texttt{none} remain separate. At the fine level, predictions are evaluated using the original six change-state labels. This multi-level reward gives partial credit to predictions that preserve the clinical direction but miss the fine-grained change state, while assigning lower rewards to direction reversals and omitted changes.

\begin{table*}[t]
\centering
\small
\setlength{\tabcolsep}{2.5pt}
\renewcommand{\arraystretch}{1.0}
\begin{tabular*}{\textwidth}{@{\extracolsep{\fill}}clcccccccccc@{}}
\toprule
\multicolumn{1}{c}{\multirow{2}{*}{\textbf{Input}}} & \multicolumn{1}{c}{\multirow{2}{*}{\textbf{Method}}} & \multirow{2}{*}{\textbf{Venue}}
& \multicolumn{6}{c}{\textbf{NLG Metrics} $\uparrow$}
& \multicolumn{3}{c}{\textbf{CE Metrics} $\uparrow$} \\
\cmidrule(lr){4-9} \cmidrule(lr){10-12}
 & &
 & B-1 & B-2 & B-3 & B-4 & R-L & MTR
 & P & R & F1 \\
\midrule

\multirow{8}{*}{\begin{tabular}{@{}c@{}}Single\\Image\end{tabular}}
& R2Gen          & EMNLP'20    & 0.308 & 0.190 & 0.126 & 0.089 & 0.266 & 0.124 & 0.457 & 0.290 & 0.355 \\
& R2GenCMN       & ACL'21      & 0.328 & 0.203 & 0.135 & 0.096 & 0.273 & 0.133 & 0.512 & 0.359 & 0.422 \\
& R2GenGPT       & Meta-Rad'23 & 0.396 & 0.243 & 0.161 & 0.108 & 0.260 & 0.155 & 0.495 & 0.385 & 0.404 \\
& PromptMRG      & AAAI'24     & 0.384 & 0.229 & 0.149 & 0.104 & 0.261 & 0.146 & 0.517 & 0.453 & 0.439 \\
& EKAGen         & CVPR'24     & 0.405 & 0.245 & 0.156 & 0.105 & 0.264 & 0.154 & 0.472 & 0.404 & 0.411 \\
& GMoD           & MICCAI'24   & 0.378 & 0.234 & 0.155 & 0.107 & 0.276 & 0.162 & 0.496 & 0.429 & 0.460 \\
& RADAR          & ACL'25      & 0.412 & 0.242 & 0.162 & 0.114 & 0.257 & 0.155 & 0.448 & 0.436 & 0.417 \\
& MedRAX          & ICML'25      & 0.388 & 0.235 & 0.151 & 0.102 & 0.259 & 0.151 & 0.524 & 0.507 & 0.515 \\

\midrule

\multirow{12}{*}{\begin{tabular}{@{}c@{}}Longitudinal\\Images\end{tabular}}
& Prefilling     & MICCAI'23   & 0.343 & 0.210 & 0.141 & 0.100 & 0.274 & 0.137 & 0.506 & 0.364 & 0.423 \\
& HERGen         & ECCV'24     & 0.389 & 0.242 & 0.163 & 0.117 & 0.282 & 0.155 & 0.421 & 0.289 & 0.295 \\
& STREAM         & TMI'25      & 0.394 & 0.237 & 0.144 & 0.104 & 0.261 & 0.143 & 0.472 & 0.428 & 0.411 \\
& MLRG           & CVPR'25     & 0.416 & 0.252 & 0.157 & 0.114 & 0.264 & 0.158 & 0.507 & 0.425 & 0.418 \\
& LLM-RG4        & AAAI'25     & 0.417 & 0.240 & 0.155 & 0.115 & 0.257 & 0.147 & 0.498 & 0.441 & 0.436 \\
& HC-LLM         & AAAI'25     & 0.404 & 0.247 & 0.164 & 0.116 & 0.271 & 0.163 & 0.488 & 0.415 & 0.448 \\
& Diff-RRG       & MICCAI'25   & 0.405 & 0.251 & 0.169 & 0.120 & 0.276 & 0.164 & 0.528 & 0.430 & 0.474 \\
& PriorRG        & AAAI'26     & 0.369 & 0.261 & \underline{0.199} & \underline{0.159} & \textbf{0.337} & 0.170 & \underline{0.576} & 0.452 & 0.507 \\
& MARE           & AAAI'26     & 0.409 & \underline{0.265} & 0.184 & 0.133 & 0.291 & 0.161 & 0.433 & 0.378 & 0.375 \\
& BiOTPrompt     & CVPR'26     & 0.397 & 0.253 & 0.174 & 0.126 & 0.285 & 0.155 & 0.471 & 0.424 & 0.417 \\
& TIM            & CVPR'26     & \underline{0.430} & \underline{0.265} & 0.179 & 0.124 & 0.287 & \underline{0.185} & 0.563 & \underline{0.505} & \underline{0.511} \\

\cmidrule(lr){2-12}
& \textbf{Ours}  & -           & \textbf{0.466} & \textbf{0.318} & \textbf{0.235} & \textbf{0.183} & \underline{0.335} & \textbf{0.195} & \textbf{0.581} & \textbf{0.665} & \textbf{0.620} \\

\bottomrule
\end{tabular*}
\caption{Comparison with single-image and longitudinal RRG methods on Longitudinal-MIMIC. B-$n$, R-L, and MTR denote BLEU-$n$, ROUGE-L, and METEOR. P, R, and F1 are the micro-averaged CheXbert precision, recall, and F1. \textbf{Bold} and \underline{underline} indicate the best and second-best result per metric.}
\label{tab:sota_main}
\end{table*}

\subsection{Verified Structured Clinical State}
\subsubsection{Consistency Gate}
Role-specialized clinical decision agents may yield mutually inconsistent predictions,
as each agent addresses a distinct clinical objective. We therefore introduce a
deterministic Consistency Gate that enforces two coherence properties between the
diagnosis and change states, leaving all other predictions unchanged. First, a diagnosis
of \texttt{NEG} is incompatible with \texttt{new} and \texttt{increased}, which the gate
maps to \texttt{decreased} if the finding is present in the prior report and to \texttt{none} otherwise, whereas a diagnosis of \texttt{POS} is incompatible
with \texttt{resolved}, which is replaced by \texttt{decreased} to preserve the direction
of change. Second, a finding diagnosed as \texttt{POS} must carry a temporal status,
since the five change states are exhaustive for a present finding; the gate therefore
completes a remaining \texttt{none} to \texttt{stable}, the only state that asserts no
change. The corrected change state is denoted by $\bar m_{t,d}$.

\subsubsection{Structured Clinical State}
\label{sec:scs}
We aggregate the outputs of the clinical decision agents after consistency correction into a Structured Clinical State (SCS), denoted by $\mathcal S_t$. For each finding $d$, we define
\begin{equation}
\mathbf z_{t,d}
=
\left(
\hat s_{t,d},
\hat{\mathbf a}_{t,d},
\bar m_{t,d}
\right),
\qquad
\mathcal S_t
=
\left\{
\mathbf z_{t,d}
\right\}_{d\in\mathcal D}.
\label{eq:scs}
\end{equation}
Here, $\hat{\mathbf a}_{t,d}=\bot$ when attribute estimation is unavailable or not applicable, whereas $\bar m_{t,d}=\mathtt{none}$ indicates that no explicit temporal change is assigned. The SCS preserves explicit disease-wise evidence for report generation and supports tracing errors to the corresponding clinical decision.

\subsection{State-Grounded Report Generation and Validation}
\subsubsection{Report Generation}
A retrieval module selects the top-$K$ training reports using IDF-weighted cosine similarity between binary disease signatures derived from $\mathcal S_t$, yielding the exemplar set $\mathcal R_t$. 
A frozen image-to-report model $\mathcal B$ generates a current-study draft $b_t$, and a frozen instruction-tuned Writer $\mathcal W$ produces the report conditioned on the SCS, the base draft, the retrieved exemplars, and the clinical context:
\begin{equation}
    b_t=\mathcal B(X_t,c_t),
    \qquad
    r_t^{\mathrm W}
    =\mathcal W(\mathcal S_t,b_t,\mathcal R_t,c_t).
    \label{eq:writer}
\end{equation}
The SCS provides the primary clinical content, while the base draft and retrieved reports supply complementary image details and stylistic guidance.

\subsubsection{Validation Agent}
Although the SCS explicitly specifies the clinical content to be reported, the generation process may still omit committed findings or express them inconsistently~\cite{nishino2022factual}. The Validation Agent therefore compares the draft report $r_t^{\mathrm W}$ with $\mathcal S_t$ and performs targeted local edits. It first extracts the diagnosis states from the draft using a CheXbert labeler. Positive findings missing from the report are added, whereas positive statements unsupported by the SCS are removed only when the diagnosis experts also provide weak current-image evidence. A second pass verifies that each change state is expressed for the correct finding and corrects missing or inconsistent descriptions. The resulting report is denoted by $\hat r_t$.

\begin{table*}[t]
\centering
\small
\setlength{\tabcolsep}{3pt}
\begin{tabular*}{\textwidth}{@{\extracolsep{\fill}}lccccccccc}
\toprule
\multicolumn{1}{c}{\multirow{2}{*}{\textbf{Method}}} & \multicolumn{2}{c}{\textbf{LCC} $\uparrow$} & \multicolumn{7}{c}{\textbf{ReXrank} $\uparrow$} \\
\cmidrule(lr){2-3} \cmidrule(lr){4-10}
 & LCC-C & LCC-F & 1/RadCliQ-v1 & BLEU & BERTScore & SembScore & RadGraph-F1 & RaTEScore & GREEN \\
\midrule
Prefilling  & 0.089 & 0.053 & 0.819 & 0.176 & 0.397 & 0.352 & 0.180 & 0.528 & 0.281 \\
HERGen      & 0.163 & 0.100 & 0.866 & 0.200 & 0.410 & 0.369 & 0.201 & 0.538 & 0.298 \\
STREAM      & 0.166 & 0.100 & 0.820 & 0.195 & 0.401 & 0.333 & 0.191 & 0.537 & 0.277 \\
MLRG        & 0.141 & 0.082 & 0.819 & 0.193 & 0.377 & 0.374 & 0.189 & 0.530 & 0.299 \\
LLM-RG4     & 0.185 & 0.115 & 0.925 & 0.216 & 0.428 & 0.395 & \underline{0.216} & 0.554 & 0.337 \\
HC-LLM      & 0.155 & 0.089 & 0.873 & 0.214 & 0.421 & 0.356 & 0.205 & 0.540 & 0.297 \\
Diff-RRG    & 0.187 & 0.114 & 0.900 & 0.220 & 0.426 & 0.376 & 0.211 & 0.542 & 0.322 \\
PriorRG     & 0.187 & 0.115 & \underline{1.141} & \underline{0.270} & \textbf{0.484} & \underline{0.441} & \textbf{0.270} & \underline{0.577} & \underline{0.341} \\
BiOTPrompt  & 0.154 & 0.092 & 0.907 & 0.214 & 0.427 & 0.381 & 0.212 & 0.543 & 0.315 \\
MedRAX  & \underline{0.193} & \underline{0.128} & 0.838 & 0.189 & 0.402 & 0.332 & 0.211 & 0.551 & 0.298 \\
\midrule
Ours (w/o Temporal) & 0.148 & 0.095 & 1.132 & 0.267 & 0.467 & 0.475 & 0.260 & 0.582 & 0.337 \\
\textbf{Ours} & \textbf{0.394} & \textbf{0.283} & \textbf{1.173} & \textbf{0.272} & \underline{0.483} & \textbf{0.467} & \textbf{0.270} & \textbf{0.582} & \textbf{0.342} \\
\bottomrule
\end{tabular*}
\caption{LCC and ReXrank results on Longitudinal-MIMIC. \textbf{Bold} and \underline{underline} indicate the best and second-best results per metric. The ablation variant is excluded from the ranking.}
\label{tab:main_lcc_rexrank}
\end{table*}

\begin{table}[t]
\centering
\setlength{\tabcolsep}{3.5pt} 
\begin{tabular}{lccccc}
\toprule
Variant & B-1 & MTR & CE-F1 & LCC-C & LCC-F \\
\midrule
Full model      & 0.466 & 0.195 & 0.620 & 0.394 & 0.283 \\
\midrule
w/o Attribute   & 0.454 & 0.191 & 0.615 & 0.346 & 0.250 \\
w/o Temporal    & 0.452 & 0.192 & 0.622 & 0.148 & 0.095 \\
w/o GRPO        & 0.464 & 0.195 & 0.618 & 0.389 & 0.259 \\
w/o Consistency & 0.465 & 0.195 & 0.620 & 0.389 & 0.274 \\
w/o Base draft  & 0.381 & 0.171 & 0.610 & 0.404 & 0.280 \\
w/o Validation  & 0.464 & 0.195 & 0.615 & 0.338 & 0.233 \\
w/o Diagnosis   & 0.458 & 0.193 & 0.612 & 0.386 & 0.278 \\
\bottomrule
\end{tabular}
\caption{Component ablation on Longitudinal-MIMIC.}
\label{tab:ablation}
\end{table}

\section{Experiments}
\label{sec:experiments}

\subsection{Experimental Setup}
\label{sec:setup}

\subsubsection{Dataset}
We evaluate on Longitudinal-MIMIC~\cite{zhu2023utilizing}, the longitudinal subset of MIMIC-CXR~\cite{johnson2024mimiccxrjpg}, and the standard benchmark for LRRG on which most recent longitudinal methods are compared~\cite{dong2026tim,gao2026mare,yun2025diff}. Each example pairs a current study with the patient's most recent prior study and its report, and the reference report accordingly describes both the current findings and how they have changed relative to the prior study. We follow the official train, validation, and test partitions, with $2{,}058$ studies in the test set, and evaluate against the released reference reports.

\subsubsection{Implementation Details}
All three clinical decision agents are optimized through supervised fine-tuning, and the Temporal Change Agent is subsequently refined using the Progression-Aware GRPO objective in Eq.~\ref{eq:temporal_reward}. Model checkpoints are selected according to validation performance. For report generation, a frozen PriorRG model~\cite{liu2026priorrg} provides a current-study draft, and we set $K{=}3$ for the retrieved exemplars. A frozen instruction-tuned $27$B LLM performs both report writing and validation-based editing.

\subsubsection{Baselines}
We compare against two groups of methods. Single-image RRG maps one chest X-ray to a report without temporal context and comprises R2Gen~\cite{chen2020r2gen}, R2GenCMN~\cite{chen2021r2gencmn}, R2GenGPT~\cite{wang2023r2gengpt}, PromptMRG~\cite{jin2024promptmrg}, EKAGen~\cite{bu2024instance}, GMoD~\cite{xiang2024gmod}, RADAR~\cite{hou2025radar}, and MedRAX~\cite{fallahpour2025medrax}. Longitudinal RRG conditions on prior images, reports, or clinical context and comprises Prefilling~\cite{zhu2023utilizing}, HERGen~\cite{wang2024hergen}, STREAM~\cite{yang2025spatio}, MLRG~\cite{liu2025enhanced}, LLM-RG4~\cite{wang2025llmrg4}, HC-LLM~\cite{liu2025hc}, Diff-RRG~\cite{yun2025diff}, PriorRG~\cite{liu2026priorrg}, MARE~\cite{gao2026mare}, BiOTPrompt~\cite{liu2026biotprompt}, and TIM~\cite{dong2026tim}. 

For Table~\ref{tab:sota_main}, we use the Longitudinal-MIMIC results reported in the original baseline papers when available and otherwise adopt the corresponding reimplementation results from TIM~\cite{dong2026tim}. If neither source provides the results, we train and evaluate the method using its official code. Table~\ref{tab:main_lcc_rexrank} requires access to the generated reports, so we retrain under the same protocol every baseline whose code is publicly released; MARE and TIM are therefore excluded, whereas MedRAX, although single-image, is retained as a representative agentic method.

\subsection{Evaluation Metrics}
\label{sec:metrics}

\subsubsection{Report Quality and Clinical Efficacy}
We report standard natural language generation (NLG) metrics, namely BLEU-1 to BLEU-4~\cite{papineni2002bleu}, ROUGE-L~\cite{lin2004rouge}, and METEOR~\cite{banerjee2005meteor}. We measure clinical efficacy (CE) by using CheXbert~\cite{smit2020chexbert} on the generated and the reference report and comparing the resulting $14$-finding labels, reporting micro-averaged precision, recall, and F1. To assess clinical quality beyond surface overlap, we further adopt the ReXrank~\cite{zhang2025rexrank} suite, which measures lexical and semantic similarity through BERTScore~\cite{zhangbertscore} and SembScore~\cite{smit2020chexbert} and clinically aware quality through RadGraph-F1~\cite{jain1radgraph}, $1/$RadCliQ-v1~\cite{yu2023evaluating}, RaTEScore~\cite{zhao2024ratescore}, and GREEN~\cite{ostmeier2024green}, together with a corpus-level BLEU.

\subsubsection{Longitudinal Change Concordance}
To directly evaluate longitudinal correctness, we report LCC, which measures whether a generated report preserves the finding-specific temporal changes stated in the reference. An instruction-tuned Gemma-4-31B extracts change propositions, each consisting of a finding and one of five labels: \texttt{new}, \texttt{increased}, \texttt{decreased}, \texttt{resolved}, or \texttt{stable}. This label set omits \texttt{none}, since a proposition is extracted only where a report states a change. A deterministic scorer matches propositions from the generated and reference reports at two levels. Coarse matching (LCC-C) requires agreement on the finding and its clinical direction, collapsing \texttt{new} into \texttt{increased} and \texttt{resolved} into \texttt{decreased}, whereas fine matching (LCC-F) requires the exact change label. Macro-F1 is computed over the reference-stated changes at each level. Unlike general semantic and clinically aware report-level metrics, this reference-anchored evaluation penalizes omitted changes and distinguishes direction-preserving errors from direction reversals.

\begin{figure*}[t!]
  \centering
  \includegraphics[width=1\linewidth]{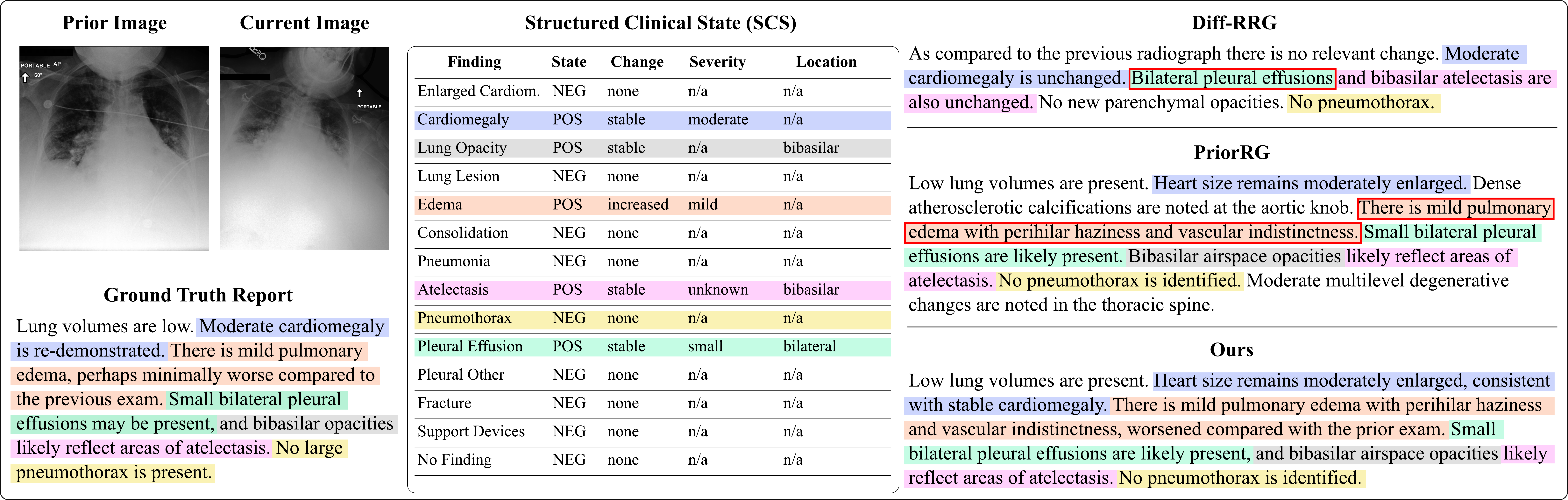}
  \caption{Qualitative comparison on Longitudinal-MIMIC. Colors indicate disease states matched to the reference report, while red boxes denote incomplete states. STRIVE captures all findings and their detailed clinical states.}
  \label{fig:qualitative}
\end{figure*}

\subsection{Main Results and Discussion}
\label{sec:main_results}

\subsubsection{Comparison with the State of the Art}

Table~\ref{tab:sota_main} compares STRIVE with single-image and longitudinal RRG methods on Longitudinal-MIMIC. STRIVE achieves the best performance on all NLG metrics except ROUGE-L, where it ranks second, and on all three CE metrics. The larger gains in CE than in NLG suggest that the improvement primarily reflects clinical correctness rather than surface-level similarity, supported by explicitly representing diagnosis, attributes, and progression in the SCS before report generation. Moreover, recall exceeding precision ($0.665$ vs. $0.581$) suggests that the framework recovers under-expressed clinical findings, while its leading precision indicates that this broader coverage does not introduce unsupported findings.

Table~\ref{tab:main_lcc_rexrank} evaluates report quality with the ReXrank suite, which measures lexical, semantic, and clinically aware agreement with the reference. STRIVE obtains the best result on six of its seven metrics, namely $1/$RadCliQ-v1 ($1.173$), BLEU ($0.272$), SembScore ($0.467$), RadGraph-F1 ($0.270$), RaTEScore ($0.582$), and GREEN ($0.342$), and ranks second on BERTScore. Since these metrics reward factual and relational agreement rather than n-gram overlap, organizing clinical evidence before generation improves not only what the report states but how faithfully it does so.

\subsubsection{Longitudinal Correctness}
Table~\ref{tab:main_lcc_rexrank} evaluates longitudinal correctness using LCC. STRIVE achieves an LCC-C of $0.394$ and an LCC-F of $0.283$, more than twice the strongest baseline scores of $0.193$ and $0.128$, respectively. These gains indicate improved accuracy in both clinical direction and fine-grained change states, whereas all baselines remain below $0.20$ despite competitive ReXrank performance. Removing the Temporal Change Agent collapses LCC while leaving ReXrank performance largely unchanged, showing that the LCC improvement is attributable to explicit temporal change reasoning rather than to general report quality.

\subsection{Ablation Study}
\label{sec:ablation}
Table~\ref{tab:ablation} presents ablation results on Longitudinal-MIMIC, where each variant removes or replaces a single component while keeping the remaining pipeline and decoding settings unchanged. The ablation results yield three insights. First, the Temporal Change Agent is the primary contributor to longitudinal correctness. Removing it sharply reduces LCC-C from $0.394$ to $0.148$ and LCC-F from $0.283$ to $0.095$, while CE-F1 remains essentially unchanged. Without the GRPO stage, LCC-F decreases to $0.259$, indicating that the proposed reward improves fine-grained discrimination among change states. Second, the two verification stages provide complementary benefits. Removing the Validation Agent reduces CE-F1 from $0.620$ to $0.615$, LCC-C from $0.394$ to $0.338$, and LCC-F from $0.283$ to $0.233$, demonstrating its role in recovering findings and change states omitted during report generation. In contrast, removing the Consistency Gate leaves CE-F1 unchanged at $0.620$ but lowers LCC-C from $0.394$ to $0.389$ and LCC-F from $0.283$ to $0.274$. This indicates that the gate specifically improves the consistency of longitudinal change predictions without altering finding-presence accuracy. Third, the Diagnosis and Attribute Agents, together with the base draft, contribute complementary information. Replacing the Diagnosis Agent with rule-based voting over the same seven experts reduces CE-F1 to $0.612$, demonstrating the benefit of learned aggregation across heterogeneous experts. Removing the Attribute Agent lowers both NLG and CE performance, indicating the importance of severity and location information. In contrast, substituting the prior report for the base draft causes the largest NLG decline, with BLEU-1 decreasing from $0.466$ to $0.381$, but only a modest reduction in CE-F1. This suggests that the base draft primarily supports report realization, while the clinical decision agents determine the core clinical content.

\subsection{Qualitative Analysis}
\label{sec:qualitative}
Figure~\ref{fig:qualitative} compares STRIVE with Diff-RRG and PriorRG on a representative longitudinal case, together with the SCS committed before generation. Both baselines recognize several findings from the reference report but leave their clinical state incomplete. Diff-RRG omits edema and lung opacity and reports pleural effusion without its severity, whereas PriorRG recognizes the pulmonary edema but not its interval increase. The SCS instead captures every finding stated in the reference together with its severity, location, and change, and STRIVE realizes all of these fields in the final report. The generated report therefore agrees with the reference not only on which abnormalities are present but also on how they are characterized and how they have changed.

\section{Conclusion}

In this work, we proposed STRIVE, a multi-agent framework for LRRG. Rather than handling all clinical reasoning tasks within a single model, it assigns diagnosis, clinical attribute estimation, and temporal change reasoning to specialized agents. Furthermore, Progression-Aware GRPO improves directional change modeling, while the Consistency Gate and Validation Agent ensure consistency between structured clinical evidence and the generated report. On Longitudinal-MIMIC, STRIVE improves report quality, clinical efficacy, and LCC, more than doubling the LCC of the strongest baseline.













\bibliography{aaai2027}


\clearpage
\appendix

\setcounter{secnumdepth}{2}
\setcounter{section}{0}
\setcounter{subsection}{0}
\renewcommand{\thesection}{S\arabic{section}}
\renewcommand{\thesubsection}{\thesection.\arabic{subsection}}

\setcounter{figure}{0}
\renewcommand{\thefigure}{S\arabic{figure}}

\setcounter{table}{0}
\renewcommand{\thetable}{S\arabic{table}}

\setcounter{equation}{0}
\renewcommand{\theequation}{S\arabic{equation}}

\section*{Supplementary Material}


\lstset{%
	basicstyle={\footnotesize\ttfamily},
	numbers=left,numberstyle=\footnotesize,xleftmargin=2em,
	aboveskip=0pt,belowskip=0pt,%
	showstringspaces=false,tabsize=2,breaklines=true}
\lstdefinestyle{promptbox}{%
	basicstyle=\footnotesize\ttfamily,
	numbers=none,
	frame=tb,
	framerule=0.5pt,
	framesep=3pt,
	xleftmargin=0pt,
	xrightmargin=0pt,
	aboveskip=4pt,
	belowskip=4pt,
	showstringspaces=false,
	keepspaces=true,
	tabsize=2,
	breaklines=true,
	breakindent=0pt,
	breakautoindent=false,
	columns=fullflexible}
\pdfinfo{
/TemplateVersion (2027.1)
}

\section{Dataset}
\label{app:dataset}
\subsection{Dataset Construction}
Longitudinal-MIMIC~\cite{zhu2023utilizing} is built from the $26{,}625$ MIMIC-CXR~\cite{johnson2024mimiccxrjpg} patients with two visits. Studies whose report has no findings section are discarded, and the remainder is partitioned with the official MIMIC-CXR patient-level split, so no patient appears in more than one partition. Each example pairs the current study with the patient's most recent prior study and its report. Table~\ref{tab:app_split} lists the resulting statistics.

\subsection{Temporal Change Statistics}
To characterize how much temporal content the reports actually carry, we apply the LCC change-statement extractor of Section~\ref{app:lcc} to the reference reports of all three partitions. Two properties motivate the LCC evaluation. First, temporal statements are common but far from universal: a comparison is stated in $55.0\%$ of the training reports, $56.2\%$ of the validation reports, and $67.5\%$ of the test reports, at a density of $1.27$ to $1.59$ statements per report. Second, the change labels are strongly imbalanced in every partition. On the test partition they are dominated by \texttt{stable} ($2{,}219$), followed by \texttt{increased} ($410$), \texttt{decreased} ($384$), \texttt{new} ($175$), and \texttt{resolved} ($87$), so the rarest state accounts for under $3\%$ of the statements. We therefore compute LCC as a macro-F1 over the change labels, so that the rare directional states are not absorbed by the dominant \texttt{stable} class.

\section{Implementation Details}
\label{app:impl}
\subsection{Clinical Decision Agents}
\paragraph{Training Configuration}
The Diagnosis and Temporal Change Agents use Gemma-4-E4B-it; the Attribute Agent uses MedGemma-1.5-4B-it. All backbones are frozen and adapted with completion-only bfloat16 LoRA ($r{=}16$, $\alpha{=}32$, dropout $0.05$). Checkpoints are selected on the patient-disjoint validation partition. Table~\ref{tab:app_hparam} gives the optimization settings. Training uses two NVIDIA RTX A6000 GPUs with $48$\,GB each.

\paragraph{Chest X-ray Expert Pool for the Diagnosis Agent}
\label{app:experts}
The Diagnosis Agent aggregates three classification experts (ConvNeXt~\cite{liu2022convnet}, RAD-DINO~\cite{perez2025exploring}, and CheXFound~\cite{yang2025chest}) and four generative experts (MedGemma~\cite{sellergren2025medgemma}, PriorRG~\cite{liu2026priorrg}, CheXagent~\cite{chen2024chexagent}, and MAIRA-2~\cite{bannur2024maira}). All seven experts are frozen while the Diagnosis Agent is trained and evaluated. The classification experts predict continuous scores for the $14$ CheXpert findings, which also provide the per-finding probabilities $\mathbf p_{t-1}$ and $\mathbf p_t$ for the Temporal Change Agent. A frozen CheXbert labeler~\cite{smit2020chexbert} maps the reports of the generative experts to categorical finding states, and all evidence is serialized in a fixed finding order.

\paragraph{Progression-Aware GRPO for Temporal Change Agent}
The Temporal Change Agent is first trained with SFT and subsequently optimized with GRPO using the same LoRA configuration. For each prompt, the policy samples a group of $G{=}6$ completions at temperature $1.0$. Group-relative advantages are computed using the progression-aware reward defined in the main paper. Because the reward is computed directly from the sampled change states and the report-derived target, GRPO does not require a separately learned reward model.

\begin{table}[t]
\centering
\setlength{\tabcolsep}{4pt}
\begin{tabular}{lrrr}
\toprule
 & Train & Val & Test \\
\midrule
Patients & $26{,}156$ & $203$ & $266$ \\
Studies  & $92{,}374$ & $737$ & $2{,}058$ \\
\midrule
Reports with change statements & $50{,}849$ & $414$ & $1{,}390$ \\
Finding-level change statements   & $117{,}040$ & $954$ & $3{,}275$ \\
\bottomrule
\end{tabular}
\caption{Longitudinal-MIMIC partitions. Each study is paired with the patient's most recent prior study, and the data are split at the patient level to prevent patient overlap across partitions.}
\label{tab:app_split}
\end{table}

\begin{table}[t]
\centering
\setlength{\tabcolsep}{3pt}
\begin{tabular}{lcccc}
\toprule
 & Diag. & Attr. & \multicolumn{2}{c}{Temporal} \\
\cmidrule(lr){4-5}
 & SFT & SFT & SFT & GRPO \\
\midrule
Backbone          & Gemma & MedGemma & Gemma & SFT ckpt \\
LoRA $r$/$\alpha$ & 16/32 & 16/32 & 16/32 & 16/32 \\
Optimizer         & AdamW & AdamW & AdamW & AdamW \\
Learning rate     & $1e^{-4}$ & $1e^{-4}$ & $1e^{-4}$ & $1e^{-5}$ \\
Batch size        & $4$ & $4$ & $4$ & $4$ \\
Precision         & bf16 & bf16 & bf16 & bf16 \\
\bottomrule
\end{tabular}
\caption{Optimization settings of the clinical decision agents. GRPO is initialized from the corresponding SFT checkpoint.}
\label{tab:app_hparam}
\end{table}

\subsection{Report Generation}
\paragraph{Base Draft Generation}
The base draft $b_t$ is produced by a frozen PriorRG~\cite{liu2026priorrg} image-to-report model, which supplies current-image detail such as devices and measurements. For retrieval, the positive findings of the Structured Clinical State (SCS) are converted into a binary CheXpert-$14$ signature and matched against the signatures of the training reports under an IDF-weighted cosine similarity, where the inverse document frequency of finding $d$ is $\log\!\big((N{+}1)/(n_d{+}1)\big)+1$ for a corpus of $N$ reports containing finding $d$ in $n_d$ of them. Weighting by the square root of the IDF on both sides makes the similarity favor exemplars whose rare findings match and penalize exemplars that carry extra findings. The top $K{=}3$ reports are injected as style exemplars.

\paragraph{Writer}
The Writer is a frozen instruction-tuned Qwen3.6-27B decoded greedily. Its prompt presents the SCS as the authoritative clinical content, the base draft as a source of image detail, and the retrieved reports as style references only, and it instructs the model to resolve any conflict in favor of the SCS.

\subsection{Validation Agent}
The Validation Agent uses the same frozen $27$B model and greedy decoding as the Writer. Given a draft report, it first applies the frozen CheXbert labeler and compares the resulting 14-finding state vector with the SCS. Findings marked as positive in the SCS but omitted from the draft are inserted. Conversely, positive statements unsupported by the SCS are removed only when the diagnosis experts assign the corresponding finding a current-image probability below 0.3, thereby preventing deletion based solely on discrepancies in wording. In a second pass, the agent verifies whether each finalized temporal change state is expressed for the correct finding and locally revises the corresponding sentence when an inconsistency is detected, rather than regenerating the entire report. The validation procedure is limited to two passes.

\section{Evaluation Metrics}
\label{app:metrics}

\subsection{NLG Metrics}
\label{app:nlg}
These three metrics measure surface agreement with the reference wording.
\begin{itemize}
\item \textbf{BLEU-$n$ (B-$n$)}~\cite{papineni2002bleu}: the geometric mean of modified $n$-gram precisions up to order $n$, with a brevity penalty. It is a corpus-level statistic, since the $n$-gram counts of all $2{,}058$ studies are pooled before the precision is formed, and being precision-based it penalizes content the reference does not contain.
\item \textbf{ROUGE-L (R-L)}~\cite{lin2004rouge}: the F-measure over the longest common subsequence of the two reports. It does not require contiguity, so it is the more recall-oriented of the two.
\item \textbf{METEOR (MTR)}~\cite{banerjee2005meteor}: a recall-weighted harmonic mean of unigram precision and recall over an alignment that admits stem, synonym, and paraphrase matches, with a penalty for fragmented alignments.
\end{itemize}
These overlap metrics do not distinguish clinically decisive terms from stylistic words.

\subsection{CE Metrics}
CE metrics score what the report asserts rather than how it is worded.
\begin{itemize}
\item \textbf{Precision (P), Recall (R), and F1} are computed by labeling the generated and the reference report with the CheXbert labeler~\cite{smit2020chexbert}, which assigns each of $14$ thoracic findings one of \textit{present}, \textit{absent}, \textit{uncertain}, or \textit{blank}. Following the convention of the baselines we compare against, the label is binarized by treating \textit{present} as positive and the other three as negative, and the scores are micro-averaged over all finding--study pairs.

\end{itemize}
CE measures current-study finding presence, not temporal direction.

\subsection{ReXrank Metrics}
The ReXrank suite~\cite{zhang2025rexrank} contributes seven metrics that reach beyond the fourteen-label constraint of CheXbert.
\begin{itemize}
\item \textbf{BLEU-2}: the mean of the CXR-Report-Metric study-level bigram BLEU scores, computed with the suite's own preprocessing and reported separately from the corpus BLEU of Section~\ref{app:nlg}.
\item \textbf{BERTScore}~\cite{zhangbertscore}: an F-measure over greedily matched contextual token embeddings, so it credits equivalent phrasing that shares no surface tokens.
\item \textbf{SembScore}~\cite{smit2020chexbert}: the cosine similarity between the CheXbert label embeddings of the two reports, an embedding-space relaxation of the CE comparison.
\item \textbf{RadGraph-F1}~\cite{jain1radgraph}: the F1 over the clinical entities and relations RadGraph extracts from each report.
\item \textbf{$1/$RadCliQ-v1}~\cite{yu2023evaluating}: the reciprocal of a composite error score fitted by regression to radiologist error counts, reported as a reciprocal so that larger is better.
\item \textbf{RaTEScore}~\cite{zhao2024ratescore}: an entity-level similarity that weights medical entities by clinical importance and is robust to synonymy and negation.
\item \textbf{GREEN}~\cite{ostmeier2024green}: a score derived from the clinically significant errors a language model enumerates between the two reports.
\end{itemize}

\subsection{Why LCC Is Needed}
\label{app:why_lcc}

The report-level metrics above compare each generated report with the reference as a whole, allowing correctly generated findings and common report content to outweigh errors in a small number of temporal expressions. However, these expressions encode the key clinical information that distinguishes longitudinal reporting from single-study description. For example, pleural effusion has increased and pleural effusion remains stable identify the same current finding and share most of their wording, yet describe different disease trajectories. A more consequential error occurs when improved is replaced by worsened, reversing the clinical interpretation despite otherwise high report-level similarity. Temporal statements should therefore be evaluated according to their clinical significance rather than their relatively small lexical footprint.


The ablation results in the main paper support this distinction. Removing the Temporal Change Agent has little effect on ReXrank but substantially reduces LCC, indicating that whole-report similarity can remain high even when temporal reasoning deteriorates. LCC addresses this limitation by directly evaluating the temporal changes stated in the reference reports and distinguishing direction-preserving errors from reversals.

\section{Longitudinal Change Concordance}
\label{app:lcc}

\subsection{Change-statement Extraction}
LCC employs an instruction-tuned Gemma-4-31B to transform each report into a set of structured, finding-level change statements. The extracted statements are subsequently matched and aggregated using deterministic rules. The model is used solely for extracting finding-level evidence and change labels and does not directly assign the metric score. To ensure consistent processing, the same model, prompt template, and decoding configuration are applied to both the reference reports and all generated reports.

The extractor is applied once to each report using the prompt template provided below. The corresponding user message specifies the $13$ admissible CheXpert findings (all $14$ categories except \textit{No Finding}), the five admissible change labels, and the report text. Greedy decoding is used throughout. Prior to extraction, each report is screened using a deterministic pattern matcher; reports without comparison cues are assigned an empty statement list without invoking the LLM. The initial generation budget is $1{,}024$ tokens. If decoding terminates at the length limit, the budget is iteratively doubled up to $4{,}096$ tokens to prevent long reports containing multiple findings from being truncated into incomplete statement lists.

\begin{lstlisting}[style=promptbox]
[SYSTEM]
You extract longitudinal change propositions from chest radiology reports.

Return compact JSON only: a list of objects. Each object must have exactly these keys:
- finding: one of the allowed findings
- label: one of new, increased, decreased, resolved, stable
- evidence_span: the exact sentence or clause supporting the proposition
- change_cue: the exact temporal/change phrase
- mention: the finding phrase
- laterality: left, right, bilateral, or empty string
- region: upper, mid, lower, hilar, mediastinal, or empty string

Rules:
- Extract only explicit comparison/progression statements.
- Do not infer change from current findings alone.
- stable includes unchanged, stable, similar, persistent, redemonstrated, no new, no significant change.
- new means newly present/interval development.
- increased means worsened/progressed/larger/more extensive.
- decreased means improved/smaller/less extensive/resolving/partial clearing.
- resolved means absent now after previously present/no longer seen/cleared.
- Ignore non-clinical dates and comparison boilerplate unless a finding change is stated.
- If no explicit finding-level change exists, return [].
- Do not include explanations, markdown, confidence, or extra keys.

[USER]
Allowed findings: Atelectasis, Cardiomegaly, Enlarged Cardiomediastinum, Consolidation, Edema,
Lung Lesion, Lung Opacity, Pleural Effusion, Pleural Other, Pneumonia, Pneumothorax, Fracture,
Support Devices
Allowed labels: new, increased, decreased, resolved, stable

Report:
<report text>

JSON:
\end{lstlisting}

\begin{table*}[t]
\centering
\setlength{\tabcolsep}{5pt}
\begin{tabular}{lccccccccc}
\toprule
\multirow{2}{*}{\textbf{Method}} & \multicolumn{5}{c}{\textbf{Fine change label} (F1) $\uparrow$}
 & \multirow{2}{*}{\textbf{LCC-F}} & \multicolumn{2}{c}{\textbf{Direction} (F1) $\uparrow$}
 & \multirow{2}{*}{\textbf{LCC-C}} \\
\cmidrule(lr){2-6} \cmidrule(lr){8-9}
 & new & increased & decreased & resolved & stable
 & & worsened & improved & \\
\midrule
Prefilling  & 0.000 & 0.024 & 0.005 & 0.000 & 0.236 & 0.053 & 0.027 & 0.004 & 0.089 \\
HERGen      & 0.040 & 0.071 & 0.064 & 0.000 & 0.325 & 0.100 & 0.107 & 0.057 & 0.163 \\
STREAM      & 0.039 & 0.086 & 0.039 & 0.000 & 0.336 & 0.100 & 0.127 & 0.036 & 0.166 \\
MLRG        & 0.000 & 0.068 & 0.039 & 0.000 & 0.300 & 0.082 & 0.077 & 0.045 & 0.141 \\
LLM-RG4     & 0.000 & 0.057 & \underline{0.130} & \underline{0.023} & 0.368 & 0.115 & 0.066 & \underline{0.121} & 0.185 \\
HC-LLM      & 0.000 & 0.052 & 0.076 & 0.000 & 0.318 & 0.089 & 0.049 & 0.099 & 0.155 \\
Diff-RRG    & 0.000 & 0.097 & 0.066 & 0.000 & \underline{0.405} & 0.114 & 0.083 & 0.074 & 0.187 \\
PriorRG     & 0.050 & 0.068 & 0.084 & 0.000 & 0.373 & 0.115 & 0.105 & 0.085 & 0.187 \\
BiOTPrompt  & 0.000 & 0.048 & 0.035 & 0.000 & 0.376 & 0.092 & 0.054 & 0.033 & 0.154 \\
MedRAX      & \underline{0.102} & \underline{0.113} & 0.098 & 0.000 & 0.326 & \underline{0.128} & \underline{0.145} & 0.108 & \underline{0.193} \\
\midrule
\textbf{Ours} & \textbf{0.195} & \textbf{0.316} & \textbf{0.200} & \textbf{0.193} & \textbf{0.511}
 & \textbf{0.283} & \textbf{0.407} & \textbf{0.265} & \textbf{0.394} \\
\midrule
Reference $n$ & $175$ & $410$ & $384$ & $87$ & $2{,}219$ & $3{,}275$ & $585$ & $471$ & $3{,}275$ \\
\bottomrule
\end{tabular}
\caption{Per-label decomposition of LCC. Each cell is the F1 with which a system recovers the reference changes carrying that label; LCC-F and LCC-C are the macro means over the five fine labels and the three direction classes. The \texttt{stable} column is shared by both levels and is therefore not repeated in the direction block. The last row gives the number of reference changes carrying each label, which is the denominator of the corresponding recall. \textbf{Bold} and \underline{underline} mark the best and second-best system per column.}
\label{tab:app_lcc_labels}
\end{table*}

\subsection{Matching Rules}
LCC uses reference-anchored, one-to-one matching. For each statement in the reference report, it searches the generated report for an unmatched candidate statement referring to the same finding. Laterality and anatomical region are retained by the extractor but are not imposed as hard matching constraints, because LCC is designed to isolate temporal agreement from localization specificity. When multiple compatible candidates are available, they are prioritized by exact fine-grained label agreement, followed by coarse directional agreement and then finding agreement alone. Each candidate statement can be matched at most once, preventing a single generated statement from covering multiple reference changes. If no compatible candidate is found, the reference statement is assigned the predicted label \texttt{none}, thereby encoding the omission as a false negative.

This produces one predicted label per reference statement, and macro-F1 is computed over those pairs, at the fine level over the five labels and at the coarse level after collapsing \texttt{new} into \texttt{increased} and \texttt{resolved} into \texttt{decreased}. \textit{Support Devices} statements are excluded from the primary scope. Candidate-only statements are excluded from this reference-anchored primary score.

\section{In-Depth Analysis}

\subsection{Longitudinal Change Coverage}
\label{sec:change_analysis}
\begin{figure}[h]
\centering
\includegraphics[width=\linewidth]{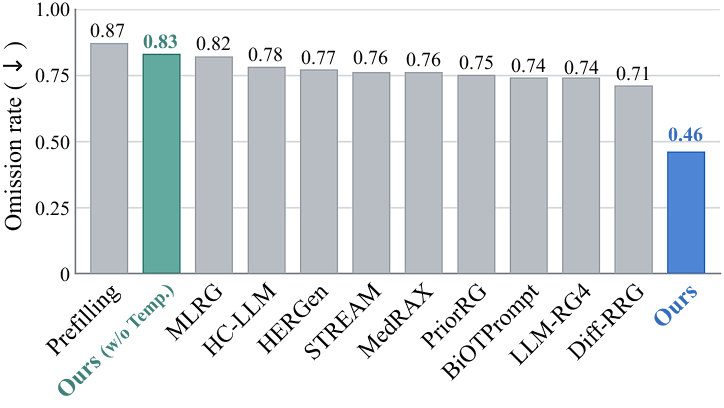}
\caption{Omission rate under the primary direction-only matching protocol, defined as the fraction of reference-stated changes absent from the generated report. STRIVE achieves the lowest omission rate, while removing the Temporal Change Agent increases it to the baseline range.}
\label{fig:omission}
\end{figure}

Figure~\ref{fig:omission} compares omission rates to assess how well each method preserves the temporal changes stated in the reference reports. The baselines omit $0.71$--$0.87$ of the reference changes, whereas STRIVE reduces the omission rate to $0.46$, corresponding to an absolute reduction of $0.25$ and a relative reduction of 35\% over the best baseline. When the Temporal Change Agent is removed, the omission rate increases from $0.46$ to $0.83$, returning to the baseline range. 

These results show that STRIVE's LCC gains arise not only from predicting the correct direction for mentioned changes but also from preserving substantially more reference-stated temporal information. Together with the improvements in coarse- and fine-grained LCC, the ablation confirms that the explicit finding-wise transition map improves both change coverage and directional accuracy by providing the Writer with committed change states that the Validation Agent can enforce.

\subsection{Per-Label Analysis of LCC}
\label{app:lcc_breakdown}
Table~\ref{tab:app_lcc_labels} decomposes LCC into per-label F1 scores, averaged over five fine-grained labels for LCC-F and three coarse-grained labels for LCC-C. STRIVE achieves the highest F1 for all five labels, confirming that its improvement is not driven solely by the dominant \texttt{stable} class. The largest gains are observed for \texttt{increased} and \texttt{resolved}; notably, nine of the ten baselines fail to recover any of the 87 reference \texttt{resolved} statements correctly. This result also motivates macro averaging: because \texttt{stable} accounts for $2{,}219$ of the $3{,}275$ reference changes, micro averaging would obscure performance on the rarer directional states, whereas macro averaging assigns equal weight to each label. This distinction is clinically important because reliable longitudinal reporting requires not only recognizing unchanged findings but also accurately capturing the rarer directional changes that indicate progression or resolution.

\begin{table*}[t]
\centering
\includegraphics[width=\textwidth]{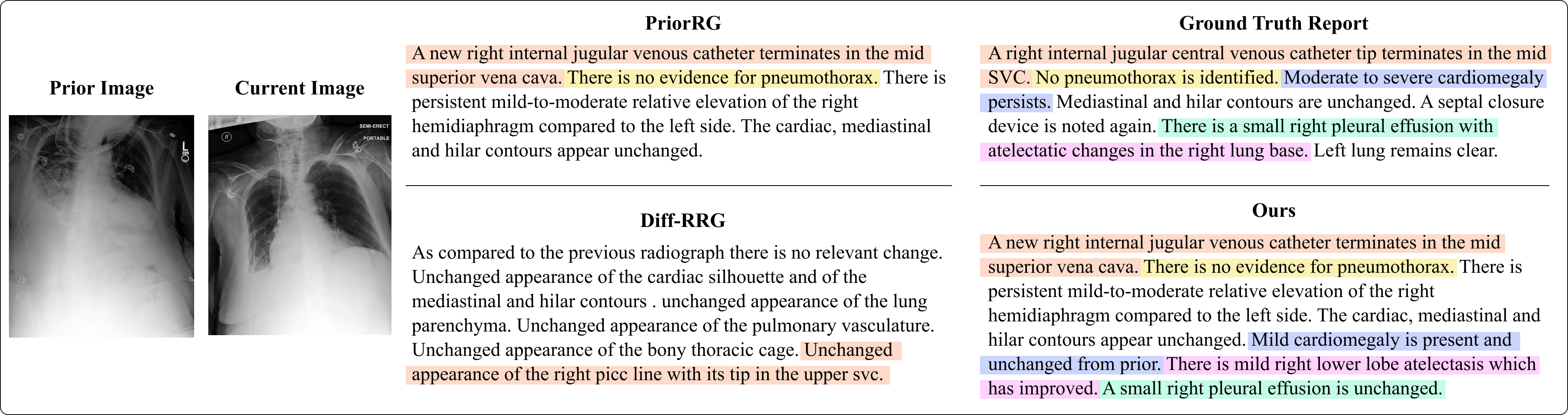}
\captionof{figure}{Additional qualitative comparison on Longitudinal-MIMIC. Colors identify finding-specific statements in the reference and generated reports: cardiomegaly (blue), atelectasis (magenta), pneumothorax (yellow), pleural effusion (green), and support devices (orange).}
\label{fig:app_qualitative}
\end{table*}

\begin{table}[t]
\centering
\small
\setlength{\tabcolsep}{1.5pt}
\begin{tabular}{@{}lcccc@{}}
\toprule
\multirow{2}{*}{\textbf{Method}} & \multicolumn{2}{c}{\textbf{LCC-C}} & \multicolumn{2}{c}{\textbf{LCC-F}} \\
\cmidrule(lr){2-3} \cmidrule(lr){4-5}
 & Value & $95\%$ CI & Value & $95\%$ CI \\
\midrule
Prefilling   & 0.089 & [0.079, 0.099] & 0.053 & [0.046, 0.060] \\
HERGen       & 0.163 & [0.146, 0.180] & 0.100 & [0.087, 0.114] \\
STREAM       & 0.166 & [0.150, 0.184] & 0.100 & [0.088, 0.113] \\
MLRG         & 0.141 & [0.126, 0.157] & 0.082 & [0.072, 0.092] \\
LLM-RG4      & 0.185 & [0.169, 0.202] & 0.115 & [0.103, 0.131] \\
HC-LLM       & 0.155 & [0.140, 0.172] & 0.089 & [0.079, 0.099] \\
Diff-RRG     & 0.187 & [0.171, 0.203] & 0.114 & [0.103, 0.124] \\
PriorRG      & 0.187 & [0.169, 0.207] & 0.115 & [0.100, 0.129] \\
BiOTPrompt   & 0.154 & [0.140, 0.169] & 0.092 & [0.083, 0.101] \\
MedRAX       & \underline{0.193} & [0.174, 0.212] & \underline{0.128} & [0.111, 0.145] \\
\midrule
\textbf{Ours} & \textbf{0.394} & [0.373, 0.417] & \textbf{0.283} & [0.255, 0.309] \\
\bottomrule
\end{tabular}
\caption{LCC scores with $95\%$ bootstrap percentile confidence intervals computed over the $2{,}058$ test study pairs using $B=2{,}000$ resamples. STRIVE's confidence interval does not overlap with that of any baseline at either granularity. \textbf{Bold} and \underline{underlined} values indicate the best and second-best point estimates, respectively.}
\label{tab:app_bootstrap}
\end{table}

\begin{table}[t]
\centering
\small
\setlength{\tabcolsep}{3pt}
\begin{tabular}{@{}lcc@{}}
\toprule
Pipeline stage & LCC-C & LCC-F \\
\midrule
Raw Temporal Agent outputs & 0.342 & 0.241 \\
\quad after Consistency Gate & 0.354 & 0.250 \\
\quad after Writer, before Validation & 0.338 & 0.233 \\
\quad after Validation & \textbf{0.394} & \textbf{0.283} \\
\bottomrule
\end{tabular}
\caption{Stage-wise LCC on the same $3{,}275$ reference changes. The first two rows score structured change states; the final two score the report before and after Validation.}
\label{tab:app_stagewise}
\end{table}

\subsection{Bootstrap Analysis of LCC Improvements}
\label{app:bootstrap}

We assessed the statistical significance of the LCC improvements using a paired nonparametric bootstrap over the $2{,}058$ test study pairs with $2{,}000$ resamples. LCC was recomputed for each resample because it is a corpus-level macro-F1 rather than an average of study-level scores. As reported in Table~\ref{tab:app_bootstrap}, STRIVE's $95\%$ confidence intervals lie entirely above those of all ten baselines at both granularities; for LCC-C, the strongest baseline has an upper bound of $0.212$, whereas STRIVE has a lower bound of $0.373$. The paired differences between STRIVE and every baseline remain positive across all resamples. Even against the strongest baseline, MedRAX, STRIVE achieves margins of $0.201$ in LCC-C ($95\%$ CI $[0.174, 0.229]$) and $0.155$ in LCC-F ($95\%$ CI $[0.126, 0.183]$), confirming that the improvements are statistically significant.

\subsection{Stage-Wise Temporal Information Analysis}
\label{app:scs_vs_report}
Table~\ref{tab:app_stagewise} traces how temporal information is preserved across the pipeline by evaluating each stage against the same $3{,}275$ reference change statements. The Temporal Change Agent achieves $0.342/0.241$ LCC-C/F, which increases to $0.354/0.250$ after the Consistency Gate. The Writer output decreases to $0.338/0.233$, indicating that some structured change states are not verbalized, whereas Validation raises the final report to $0.394/0.283$. The first two rows evaluate structured states and the last two evaluate report-extracted statements; therefore, this table provides a stage-wise pipeline trace rather than a component-removal ablation. This stage-wise trace highlights the value of explicit intermediate change states, which make temporal information loss during report generation identifiable and recoverable through validation.

\subsection{Additional Qualitative Case Analysis}
Figure~\ref{fig:app_qualitative} presents an additional case-level comparison across five finding categories highlighted in the reference report: cardiomegaly, atelectasis, pneumothorax, pleural effusion, and support devices. PriorRG~\cite{liu2026priorrg} preserves the catheter and negative pneumothorax statements but omits atelectasis and pleural effusion, whereas Diff-RRG~\cite{yun2025diff} focuses primarily on stability and line placement.

In contrast, STRIVE covers every highlighted category.

\section{Prompts}
\label{app:prompts}
This section presents the instruction and input schemas used by the language-model agents. Angle brackets mark inference-time slots, and an ellipsis marks a block repeated per finding. The Diagnosis, Attribute, and Temporal Change Agents are fine-tuned on the requested response formats, whereas the Writer and Validation Agent are frozen.

\subsection{Diagnosis Agent}
The template combines three classification probabilities and four CheXbert reads in a fixed finding order.

\begin{lstlisting}[style=promptbox]
[SYSTEM]
You are the orchestrator of a chest X-ray diagnosis system. The CURRENT study is represented by a
SINGLE chest X-ray image of a stated view (PA, AP, or LATERAL). SEVEN experts scored THIS image:
  - cn (convnext), rd (rad_dino), cf (chexfound): image models -> probability 0..1
  - mg (medgemma), pr (priorrg), cx (chexagent), mr (maira2): vision-language readers
    -> CheXbert read (1=present, 0.5=uncertain, 0=absent)
Experts disagree, and each is more reliable for some findings than others. The VIEW matters -- a
finding can be clearer or harder to see on a given view (e.g. a small effusion is clearer on
lateral; a retrocardiac opacity can be obscured on a frontal view), so calibrate confidence to what
this single view can show. For EACH of the 14 findings, integrate the seven expert reads for THIS
image and assign a STATE:
  POS = confidently present on this image
  UNC = uncertain / equivocal -- a radiologist would hedge
  NEG = absent, or not applicable
Most findings are NEG. Use POS only when the evidence is convincing; use UNC when the experts
disagree or the signal is borderline. Output ONLY a JSON object mapping each finding to
"POS"|"UNC"|"NEG". No prose.

[USER]
CURRENT study: 1 image(s).

Per-finding expert reads for EACH image/view (cn rd cf = prob 0-1; mg pr cx mr = score 0/0.5/1):
- Enlarged Cardiomediastinum: [<view>] cn=<p> rd=<p> cf=<p> mg=<s> pr=<s> cx=<s> mr=<s>
- Cardiomegaly:               [<view>] cn=<p> rd=<p> cf=<p> mg=<s> pr=<s> cx=<s> mr=<s>
  ... one line per CheXpert finding, in a fixed order ...
- No Finding:                 [<view>] cn=<p> rd=<p> cf=<p> mg=<s> pr=<s> cx=<s> mr=<s>

JSON (each finding -> "POS"|"UNC"|"NEG"):
\end{lstlisting}

\subsection{Attribute Agent}
The agent is queried once per positive finding, permits \texttt{none}, and uses an open vocabulary for anatomical modifiers.

\begin{lstlisting}[style=promptbox]
[SYSTEM]
You are an expert thoracic radiologist writing the findings section of a chest X-ray report. You are
given the current radiograph and a finding that IS present. Output ONLY how you would characterize
that finding in the report: its severity/extent, its laterality (left, right, bilateral, or a
bilateral form such as bibasilar), and its region (for example upper, mid, lower, base/basilar,
apical, hilar, perihilar, retrocardiac, costophrenic, or lingular) - using the exact words and the
natural order a radiologist writes (e.g. 'small right', 'moderate', 'left basilar', 'mild bibasilar',
'left retrocardiac', 'small bilateral', 'patchy right lower'). The lists above are examples, not a
closed vocabulary: if the radiograph calls for a different anatomical descriptor, write the one a
radiologist would use. Radiologists frequently OMIT modifiers when a finding is not prominent or not
localized; in that case output exactly 'none'. Output only the modifier words (or 'none'), nothing
else.

[USER]
<current chest X-ray image>
Finding: <finding name>. Characterization:
\end{lstlisting}

\subsection{Temporal Change Agent}
The direction token supports forward and time-reversed instances; unlisted findings map to $\mathtt{none}$.

\begin{lstlisting}[style=promptbox]
[SYSTEM]
You are an expert thoracic radiologist predicting how each chest X-ray finding CHANGED between two
studies of the same patient. You are given the DIRECTION, one study's report, the TIME GAP, and,
per finding, the prior report stance, prior and current diagnosis-model presence, prior and current
image probability with interval delta, and temporal dynamics.
Integrate these signals to determine presence, then assign new, increased, decreased, resolved, or
stable relative to the stated direction. Use the interval delta for magnitude and the time gap with
the finding's dynamics to judge plausibility.
First reason briefly inside <think>...</think>. Then output ONLY a JSON array inside
<answer>...</answer>: {"finding":<allowed finding>,"direction":<change label>}.
Report only clinically notable changes; any finding you do not list is assumed unchanged.

[USER]
DIRECTION: <PRIOR -> CURRENT | CURRENT -> PRIOR>
Allowed findings: Enlarged Cardiomediastinum, Cardiomegaly, Lung Opacity, Lung Lesion, Edema,
Consolidation, Pneumonia, Atelectasis, Pneumothorax, Pleural Effusion, Pleural Other, Fracture
TIME GAP: ~<days> days <documented or approximate>
<PRIOR- or CURRENT-STUDY REPORT>: "<report>"
PER-FINDING SIGNALS [prior report stance | prior and current diagnosis-model presence |
prior and current image probability, interval delta | temporal dynamics]:
- <finding>: prior[report<+|-|?> belief<+|-> <p>] current[belief<+|-> <p>] Delta<d> |
  <rate>, <resolution constraint>
  ... one line per allowed finding ...
Predict the notable interval changes, with direction:
\end{lstlisting}

\subsection{Writer}
The system message is abbreviated for space. The structured state controls clinical content, while retrieved reports provide style only.

\begin{lstlisting}[style=promptbox]
[SYSTEM]
You are a radiologist writing a chest-radiograph report as ONE continuous prose paragraph in natural MIMIC-CXR style (no headers, no colons, no 'Findings:'/'Impression:' labels, no preamble).

CONTENT (authoritative = the structured clinical state):
- Report EVERY positive finding so a radiologist would recognize it; hedge uncertain findings; never mention an absent finding as present. Draw devices, measurements, laterality and specific detail from the Base Draft, but the set of present findings MUST equal the positive set of the state.
- Name findings unambiguously, but use the NATURAL phrasings of the Style References.

REGISTER (write like the Style References, NOT a checklist):
- Do NOT use the phrase 'is present'. Integrate findings into flowing sentences.
- Attach a severity word (mild/moderate/severe) ONLY to a genuinely prominent (new or clearly
  worsening) finding; MOST findings need NO severity word.
- For stable or chronic findings use continuity language ('again seen', 'unchanged', 'stable',
  'persistent', 'redemonstrated', 'as before').
- Vary sentence openings; do NOT begin most sentences with 'there is'. Match the references'
  brevity, connectives, and ordering. Emulate their STYLE only; their findings are not yours.

[USER]
## Clinical Context & Patient History
<indication and history>

## Views Examined
<views>

## Structured Clinical Findings (Current Study)
REQUIRED MENTIONS (POS): <findings>
UNCERTAIN: <findings>
ABSENT: <findings>

## Attributes (severity / location / laterality per positive finding)
- <finding>: <severity> <laterality> <region>
  ... one line per positive finding ...

## Changes from Prior Study
- <finding>: <new|increased|stable|decreased|resolved>
  ... one line per finding carrying a change state ...

## Base Draft Report (image detail; carry over devices, measurements, laterality)
<base draft>

## Style References (real reports with a similar finding profile; STYLE ONLY)
1. <retrieved report>
2. <retrieved report>
3. <retrieved report>
\end{lstlisting}

\subsection{Validation Agent}
The two passes first reconcile finding presence and then insert any missing temporal states.

\begin{lstlisting}[style=promptbox]
=== PASS 1: presence ===
[SYSTEM]
You are a minimal-diff chest X-ray report editor. Apply ONLY the listed corrections; keep every
other sentence VERBATIM (same wording, order, style). Make each correction in as few words as
possible and read naturally like a radiologist. Output ONLY the edited report as one continuous
paragraph, no headers.

[USER]
## CURRENT REPORT
<draft report>

## CORRECTIONS TO APPLY (apply ONLY these; keep everything else verbatim)
- ADD <finding>: stated in the clinical state but missing from the report
- REMOVE <finding>: asserted in the report, absent from the clinical state, image probability <0.3
  ... one line per correction ...

Output ONLY the edited report as one continuous paragraph.

=== PASS 2: change ===
[SYSTEM]
You are a radiologist minimally editing a chest X-ray report to make each finding's temporal change
explicit versus the prior study, using ONE CONCISE word only. For EVERY finding listed, add the
single change word (unchanged/stable, new, increased, decreased, or resolved) DIRECTLY to that
finding's EXISTING mention as a one-word modifier. Do NOT add 'compared to the prior study', 'from
prior', or any long comparison clause -- just the single word. Examples: 'severe cardiomegaly' ->
'stable severe cardiomegaly'; 'small pleural effusion' -> 'increased small pleural effusion';
'atelectasis' -> 'new atelectasis'. Never remove the disease name or its severity. Only if a finding
is not mentioned at all, add ONE very short sentence (e.g. 'New pleural effusion.'). Keep all other
sentences verbatim. Do not add findings that are not listed. Output ONLY the edited report as one
continuous paragraph.

[USER]
## REPORT
<report after pass 1>

## FINDINGS -- make each finding's temporal change explicit
- <finding>: <change word>
  ... one line per finding whose committed change state is missing from the report ...

Edited report:
\end{lstlisting}




\end{document}